\documentclass[final]{cvpr}

\usepackage{times}
\usepackage{epsfig}
\usepackage{graphicx}
\usepackage{amsmath}
\usepackage{amssymb}
\usepackage{subcaption}
\usepackage{multirow}
\usepackage{soul}
\usepackage{textcomp}

\usepackage{listings}
\usepackage{xcolor}

\definecolor{codegreen}{rgb}{0,0.6,0}
\definecolor{codegray}{rgb}{0.5,0.5,0.5}
\definecolor{codepurple}{rgb}{0.58,0,0.82}
\definecolor{backcolour}{rgb}{0.95,0.95,0.92}

\lstdefinestyle{mystyle}{
  backgroundcolor=\color{backcolour},   commentstyle=\color{codegreen},
  keywordstyle=\color{magenta},
  numberstyle=\tiny\color{codegray},
  stringstyle=\color{codepurple},
  basicstyle=\ttfamily\footnotesize,
  breakatwhitespace=false,         
  breaklines=true,                 
  captionpos=b,                    
  keepspaces=true,                 
  numbers=left,                    
  numbersep=5pt,                  
  showspaces=false,                
  showstringspaces=false,
  showtabs=false,                  
  tabsize=2
}

\usepackage[pagebackref=true,breaklinks=true,colorlinks,bookmarks=false]{hyperref}

\def\cvprPaperID{****} % *** Enter the CVPR Paper ID here
\def\confYear{CVPR 2021}

\begin{document}

%%%%%%%%% TITLE
\title{Visual Transformers: Token-based Image Representation and Processing for Computer Vision}

\author{Bichen Wu$^1$, Chenfeng Xu$^3$, Xiaoliang Dai$^1$, Alvin Wan$^3$, Peizhao Zhang$^1$\\
Zhicheng Yan$^2$, Masayoshi, Tomizuka$^3$, Joseph Gonzalez$^3$, Kurt Keutzer$^3$, Peter Vajda$^1$ \\
$^1$ Facebook Reality Labs, $^2$ Facebook AI, $^3$ UC Berkeley \\
{\tt\small \{wbc, xiaoliangdai, stzpz, zyan3, vajdap\}@fb.com} \\
{\tt\small \{xuchenfeng, alvinwan, tomizuka, jegonzal, keutzer\}@berkeley.edu}
}

\maketitle

%%%%%%%%% ABSTRACT
\begin{abstract}
Computer vision has achieved remarkable success by (a) representing images as uniformly-arranged pixel arrays and (b) convolving highly-localized features. However, convolutions treat all image pixels equally regardless of importance; explicitly model all concepts across all images, regardless of content; and struggle to relate spatially-distant concepts. In this work, we challenge this paradigm by (a) representing images as semantic visual tokens and (b) running transformers to densely model token relationships. Critically, our \textbf{Visual Transformer} operates in a semantic token space, judiciously attending to different image parts based on context. This is in sharp contrast to pixel-space transformers that require orders-of-magnitude more compute. 
Using an advanced training recipe, our VTs significantly outperform their convolutional counterparts, raising ResNet accuracy on ImageNet top-1 by \textbf{4.6 to 7 points} while using fewer FLOPs and parameters. 
For semantic segmentation on LIP and COCO-stuff, VT-based feature pyramid networks (FPN) achieve 0.35 points higher mIoU while reducing the FPN module's FLOPs by \textbf{6.5x}.
\end{abstract}

\section{Introduction}
In computer vision, visual information is captured as arrays of pixels. These pixel arrays are then processed by convolutions, the \textit{de facto} deep learning operator for computer vision. Although this convention has produced highly successful vision models, there are critical challenges:

\begin{figure*}[!htb]
\vspace{-0.5cm}
  \centering
  \includegraphics[width=.9\linewidth]{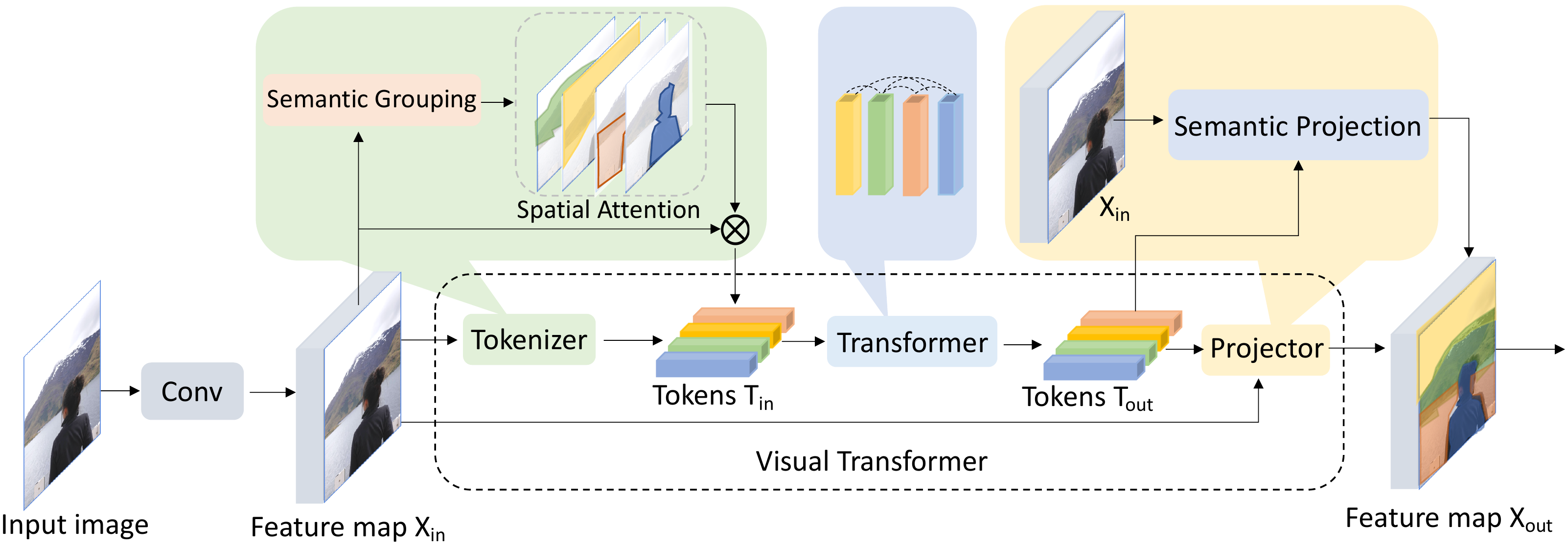}
  \caption{Diagram of a Visual Transformer (VT). For a given image, we first apply convolutional layers to extract low-level features. The output feature map is then fed to VT: First, apply a \textit{tokenizer}, grouping pixels into a small number of \textit{visual tokens}, each representing a semantic concept in the image. Second, apply transformers to model relationhips between tokens. Third, visual tokens are directly used for image classification or projected back to the feature map for semantic segmentation.
  }
  \label{fig:diagram}
\end{figure*}

1) \textbf{Not all pixels are created equal}: Image classification models should prioritize foreground objects over the background. Segmentation models should prioritize pedestrians over disproportionately large swaths of sky, road, vegetation etc. Nevertheless, convolutions uniformly process all image patches regardless of importance. This leads to spatial inefficiency in both computation and representation.

2) \textbf{Not all images have all concepts}:
Low-level features such as corners and edges exist in all natural images, so applying low-level convolutional filters to \textit{all} images is appropriate. However, high-level features such as ear shape
exist in specific images, so applying high-level filters to all images is computationally inefficient.
For example, dog features may not appear in images of flowers, vehicles, aquatic animals etc. This results in rarely-used, inapplicable filters expending a significant amount of compute. 

3) \textbf{Convolutions struggle to relate spatially-distant concepts}: Each convolutional filter is constrained to operate on a small region, but long-range interactions between semantic concepts is vital. To relate spatially-distant concepts, previous approaches increase kernel sizes, increase model depth, or adopt new operations like dilated convolutions, global pooling, and non-local attention layers. However, by working within the pixel-convolution paradigm, these approaches at best mitigate the problem, compensating for the convolution's weaknesses by adding model and computational complexity.

To overcome the above challenges, we address the root cause, the pixel-convolution paradigm, and introduce the \textit{Visual Transformer} (VT) (Figure \ref{fig:diagram}), a new paradigm to represent and process high-level concepts in images. 
Our intuition is that a sentence with a few words (or visual tokens) suffices to describe high-level concepts in an image. This motivates a departure from the fixed pixel-array representation later in the network; instead, we use spatial attention to convert the feature map into a compact set of semantic tokens.
We then feed these tokens to a transformer, a self-attention module widely used in natural language processing \cite{vaswani2017attention} to capture token interactions. 
The resulting visual tokens computed can be directly used for image-level prediction tasks (e.g., classification) or be spatially re-projected to the feature map for pixel-level prediction tasks (e.g., segmentation). Unlike convolutions, our 
VT can better handle the three challenges above: 1) judiciously allocating computation by attending to important regions, instead of treating all pixels equally; 2) encoding semantic concepts in a few visual tokens relevant to the image, instead of modeling all concepts across all images; and 
3) relating spatially-distant concepts through self-attention in token-space. 

To validate the effectiveness of VT and understanding its key components, we run controlled experiments by using VTs to replace convolutions in ResNet, a common test bed for new building blocks for image classification. We also use VTs to re-design feature-pyramid networks (FPN), a strong baseline for semantic segmentation. Our expeirments show that VTs achieve higher accuracy with lower computational cost in both tasks. For the ImageNet\cite{deng2009imagenet} benchmark, we replace the last stage of ResNet\cite{he2016deep} with VTs, reducing FLOPs of the stage by 6.9x and improving top-1 accuracy by \textbf{4.6 to 7 points}. For semantic segmentation on COCO-Stuff \cite{caesar2018cvpr} and Look-Into-Person \cite{liang2018look}, VT-based FPN achieves 0.35 points higher mIOU while reducing regular FPN module's FLOPs by 6.4x.

\section{Relationship to previous work}

\textbf{Transformers in vision models}: A notable recent and relevant trend is the adoption of transformers in vision models. Dosovitskiy \textit{et al.} propose a Vision Transformer (ViT) \cite{dosovitskiy2020image}, dividing an image into $16\times 16$ patches and feeding these patches (i.e., tokens) into a standard transformer. Although simple, this requires transformers to learn dense, repeatable patterns (e.g., textures), which convolutions are drastically more efficient at learning. The simplicity incurs an extremely high computational price: ViT requires up to 7 GPU years and 300M JFT dataset images to outperform competing convolutional variants. By contrast, we leverage the respective strengths of each operation, using convolutions for extracting low-level features  and transformers for relating high-level concepts. We further use spatial attention to focus on important regions, instead of treating each image patch equally. This yields strong performance despite orders-of-magnitude less data and training time.

Another relevant work, DETR\cite{carion2020end}, adopts transformers to simplify the hand-crafted anchor matching procedure in object detection training. Although both adopt transformers, DETR is not directly comparable to our VT given their orthogonal use cases, i.e., insights from both works could be used together in one model for compounded benefit. 

\textbf{Graph convolutions in vision models}: Our work is also related to previous efforts such as GloRe \cite{chen2019graph}, LatentGNN \cite{zhang2019latentgnn}, and \cite{liang2018symbolic} that densely relate concepts in latent space using graph convolutions. To augment convolutions, \cite{liang2018symbolic, chen2019graph, zhang2019latentgnn} adopt a procedure similar to ours: (1) extracting latent variables to represent in graph nodes (analogous to our visual tokens) (2) applying graph convolution to capture node interactions (analogous to our transformer), and (3) projecting the nodes back to the feature map. Although these approaches avoid spatial redundancy, they are susceptible to concept redundancy: the second limitation listed in the introduction. In particular, by using fixed weights that are not content-aware, the graph convolution expects a fixed semantic concept in each node, regardless of whether the concept exists in the image. By contrast, a transformer uses content-aware weights, allowing visual tokens to represent varying concepts. As a result, while graph convolutions require hundreds of nodes (128 nodes in \cite{chen2016attention}, 340 in \cite{liang2018look}, 150 in \cite{zhang2018shufflenet}) to encode potential semantic concepts, our VT uses just 16 visual tokens and attains higher accuracy. Furthermore, while modules from \cite{liang2018symbolic, chen2019graph, zhang2019latentgnn} can only be added to a pretrained network to augment convolutions, VTs can replace convolution layers to save FLOPs and parameters, and support training from scratch.

\textbf{Attention in vision models:} In addition to being used in transformers, attention is also widely used in different forms in computer vision models \cite{hu2018squeeze,hu2018gather,woo2018cbam,wu2018squeezesegv2,xu2020squeezesegv3,wang2018non,parmar2018image,hu2018relation,hu2019local,bello2019attention,zhao2020exploring,ramachandran2019stand}. Attention was first used to modulate the feature map: attention values are computed from the input and multiplied to the feature map as in \cite{woo2018cbam,hu2018squeeze,hu2018gather, wu2018squeezesegv2}. Later work \cite{xu2020squeezesegv3, su2019pixel, wang2018depth} interpret this ``modulation'' as a way to make convolution spatially adaptive and content-aware. In \cite{wang2018non}, Wang \textit{et al.} introduced non-local operators, equivalent to self-attention, to video understanding to capture the long-range interactions. However, the computational complexity of self-attention grows quadratically with the number of pixels. \cite{bello2019attention} use self-attention to augment convolutions and reduce the compute cost by using small channel sizes for attention. \cite{ramachandran2019stand, parmar2018image,cordonnier2019relationship,zhao2020exploring,hu2019local} on the other hand restrict receptive field of self-attention and use it in a convolutional manner. Starting from \cite{ramachandran2019stand}, self-attentions are used as a stand-alone building block for vision models. Our work is different from all above since we propose a novel token-transformer paradigm to replace the inefficient pixel-convolution paradigm and achieve superior performance.

\textbf{Efficient vision models:} Many recent research efforts have been focusing on building vision models to achieve better performance with lower computational cost.
Early work in this direction includes \cite{iandola2016squeezenet,xnornet-2016,gholami2018squeezenext,howard2017mobilenets,sandler2018mobilenetv2,howard2019searching,zhang2018shufflenet,ma2018shufflenet,wu2018shift}. 
Recently, people use neural architecture search \cite{wu2019fbnet,dai2019chamnet,wan2020fbnetv2,dai2020fbnetv3,tan2019efficientnet,tan2019mnasnet} to optimize network's performance within a search space that consists of existing convolution operators. 
The efforts above all seek to make the common convolutional-neural net more computationally efficient.  
In contrast, we propose a new building block that naturally eliminates the
redundant computations in the pixel-convolution paradigm. 

\section{Visual Transformer}
\label{sec:visual_transformer}

We illustrate the overall diagram of a \textit{Visual Transformer} (VT) based model in Figure \ref{fig:diagram}. First, process the input image with several convolution blocks, then feed the output feature map to VTs.
Our insight is to leverage the strengths of both convolutions and VTs: (1) early in the network, use convolutions to learn densely-distributed, low-level patterns and (2) later in the network, use VTs to learn and relate more sparsely-distributed, higher-order semantic concepts.
At the end of the network, use visual tokens for image-level prediction tasks and use the augmented feature map for pixel-level prediction tasks.

A VT module involves three steps: First, group pixels into semantic concepts, to produce a compact set of \textit{visual tokens}. Second, to model relationships between semantic concepts, apply a transformer \cite{vaswani2017attention} to these visual tokens. 
Third, project these visual tokens back to pixel-space to obtain an augmented feature map. 
Similar paradigms can be found in \cite{chen2019graph, zhang2019latentgnn,liang2018symbolic} but with one critical difference: Previous methods use hundreds of semantic concepts (termed, ``nodes''), whereas our VT uses as few as 16 visual tokens to achieve superior performance.

\subsection{Tokenizer}
\label{sec:tokenization}

Our intuition is that an image can be summarized by a few handfuls of words, or \textit{visual tokens}. This contrasts convolutions, which use hundreds of filters, and graph convolutions, which use hundreds of ``latent nodes'' to detect all possible concepts regardless of image content. To leverage this intuition, we introduce a \textit{tokenizer} module to convert feature maps into compact sets of visual tokens. Formally, we denote the input feature map by $\mathbf{X} \in \mathbb{R}^{HW \times C}$ (height $H$, width $W$, channels $C$) and visual tokens by $\mathbf{T} \in \mathbb{R}^{L \times C}$ s.t. $L \ll HW$ ($L$ represents the number of tokens).

\subsubsection{Filter-based Tokenizer}
\label{sec:static_tokenization}
A filter-based tokenizer, also adopted by \cite{zhang2019latentgnn, chen2019graph,liang2018symbolic}, utilizes convolutions to extract visual tokens. %For a given input feature map $\mathbf{X} \in \mathbb{R}^{HW \times C}$, 
For feature map $\mathbf{X}$, we map each pixel $\mathbf{X}_p \in \mathbb{R}^C$ to one of $L$ semantic groups using point-wise convolutions.
Then, within each group, we spatially pool pixels to obtain tokens $\mathbf{T}$. Formally,
\begin{equation}
\label{eqn:conv_based_token}
\begin{aligned}
    \mathbf{T} = {\underbrace{\text{\textsc{softmax}}_{_{HW}}\left(\mathbf{X}\mathbf{W}_A\right)}_{\mathbf{A} \in \mathbb{R}^{HW\times L}}}^T \mathbf{X}
    %= \mathbf{A}^T \mathbf{X}.
\end{aligned}
\end{equation}
Here, $\mathbf{W}_A \in \mathbb{R}^{C\times L}$ forms semantic groups from $\mathbf{X}$, and $\text{\textsc{softmax}}_{_{HW}}(\cdot)$ translates these activations into a spatial attention. 
Finally, $\mathbf{A}$ multiplies with $\mathbf{X}$ and computes weighted averages of pixels in $\mathbf{X}$ to make $L$ visual tokens.

However, many high-level semantic concepts are sparse and may each appear in only a few images. As a result, the fixed set of learned weights $\mathbf{W}_A$ potentially wastes computation by modeling all such high-level concepts at once. We call this a ``filter-based'' tokenizer, since it uses convolutional filters $\mathbf{W_A}$ to extract visual tokens. 

\begin{figure}[!htb]
  \centering
  \includegraphics[width=.9\linewidth]{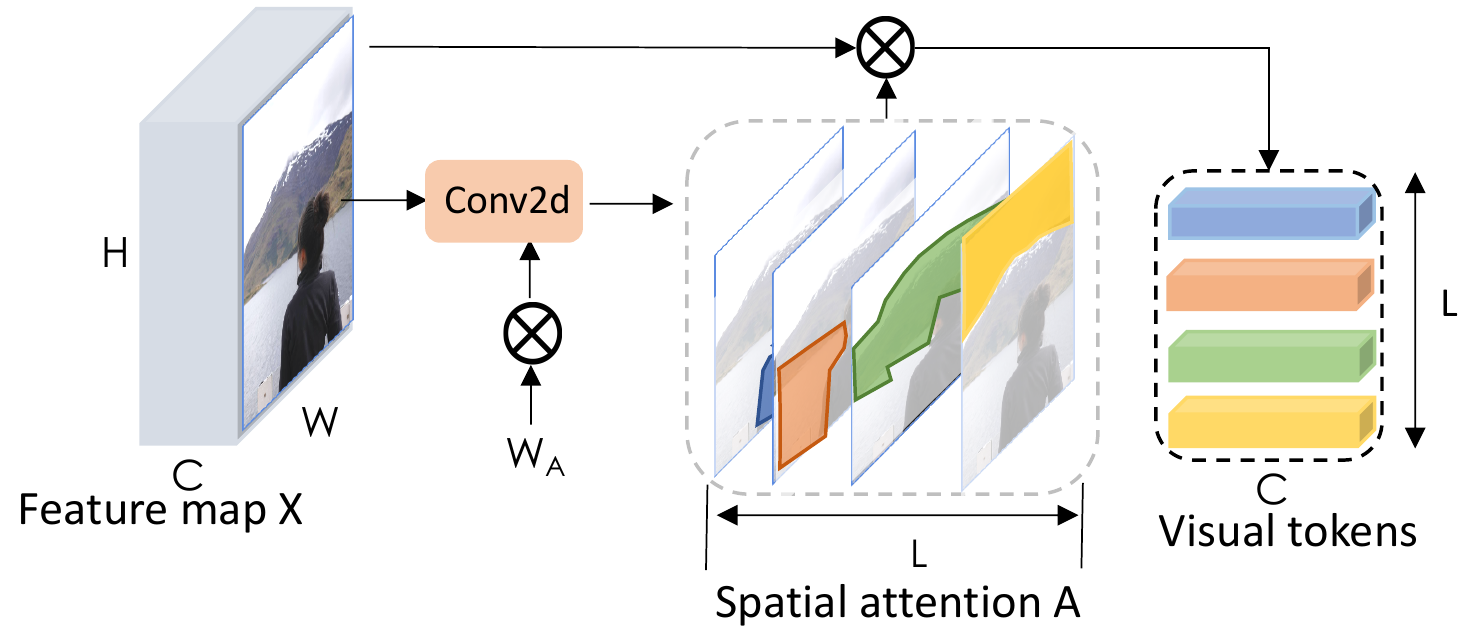}
  \caption{Filter-based tokenizer that use convolution to group pixels using a fixed convolution filter. }
  \label{fig:filter_tknzr}
\end{figure}

\subsubsection{Recurrent Tokenizer}
To remedy the limitation of filter-based tokenizers, we propose a recurrent tokenizer with weights that are dependent on previous layer's visual tokens. The intuition is to let the previous layer's tokens $\mathbf{T}_{in}$ guide the extraction of new tokens for the current layer. The name of ``recurrent tokenizer'' comes from that current tokens are computed dependent on previous ones. Formally, we define
\begin{equation}
    \begin{gathered}
        \mathbf{W}_R = \mathbf{T}_{in}\mathbf{W_{T \rightarrow R}}, \\
        \mathbf{T} = \text{\textsc{softmax}}_{_{HW}}\left(\mathbf{X}\mathbf{W}_R\right)^T \mathbf{X},
    \end{gathered}
    \label{eqn:cluster_based_token}
\end{equation}
where $\mathbf{W}_{T \rightarrow R} \in \mathbb{R}^{C \times C}$. In this way, the VT can incrementally refine the set of visual tokens, conditioned on previously-processed concepts. In practice, we apply recurrent tokenizers starting from the second VT, since it requires tokens from a previous VT.  % TODO: explain use of 'recurrent' name

\begin{figure}[!htb]
  \centering
  \includegraphics[width=.9\linewidth]{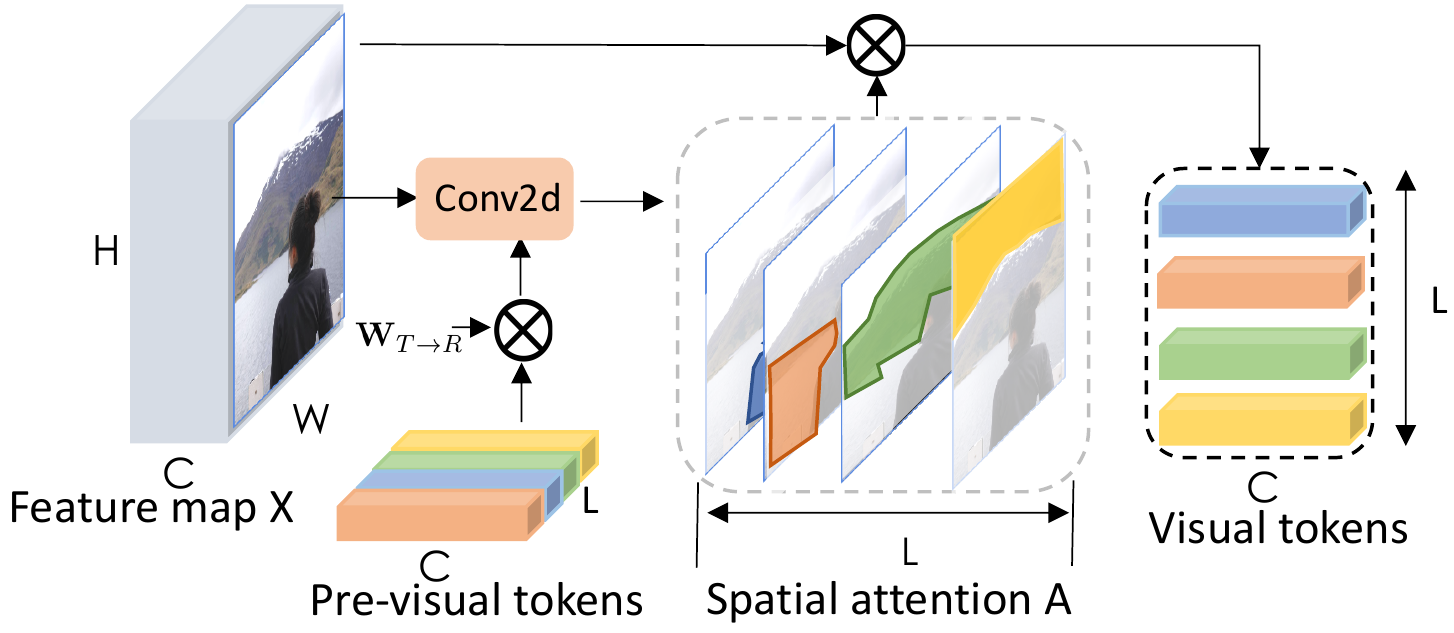}
  \caption{Recurrent tokenizer that uses previous tokens to guide the token extraction in the current VT module.}
  \label{fig:refine_tknzr}
\end{figure}

\subsection{Transformer}
\label{sec:transformer}

After tokenization, we then need to model interactions between these visual tokens. Previous works \cite{chen2019graph,zhang2019latentgnn,liang2018symbolic} use graph convolutions to relate concepts. However, these operations use fixed weights during inference, meaning each token (or ``node'') is bound to a specific concept, therefore graph convolutions waste computation by modeling all high-level concepts, even those that only appear in few images. To address this, we adopt transformers \cite{vaswani2017attention}, which use input-dependent weights by design. Due to this, transformers support visual tokens with variable meaning, covering more possible concepts with fewer tokens.

We employ a standard transformer with minor changes:
\begin{equation}
\label{eqn:self_attention}
        \mathbf{T}_{out}' = \mathbf{T}_{in} + \text{\textsc{softmax}}_{_L} \left((\mathbf{T}_{in}\mathbf{K}) (\mathbf{T}_{in}\mathbf{Q})^T\right) \mathbf{T}_{in},
\end{equation}
\begin{equation}
\label{eqn:ffn}
        \mathbf{T}_{out} = \mathbf{T}_{out}' + \sigma(\mathbf{T}_{out}'\mathbf{F}_1)\mathbf{F}_2,
\end{equation}
where $\mathbf{T}_{in}, \mathbf{T}_{out}', \mathbf{T}_{out} \in \mathbb{R}^{L\times C}$ are the visual tokens. 
Different from graph convolution, in a transformer, weights between tokens are input-dependent and computed as a key-query product: $(\mathbf{T}_{in}\mathbf{K}) (\mathbf{T}_{in}\mathbf{Q})^T \in \mathbb{R}^{L\times L}$. 
This allows us to use as few as 16 visual tokens, in contrast to hundreds of analogous nodes for graph-convolution approaches \cite{chen2019graph,zhang2019latentgnn,liang2018symbolic}. 
After the self-attention, we use a non-linearity and two pointwise convolutions in Equation (\ref{eqn:ffn}), where $\mathbf{F}_1, \mathbf{F}_2 \in \mathbb{R}^{C\times C}$ are weights, $\sigma(\cdot)$ is the ReLU function. 

\subsection{Projector}
\label{sec:projector}
Many vision tasks require pixel-level details, but such details are not preserved in visual tokens. 
Therefore, we fuse the transformer's output with the feature map to refine the feature map's pixel-array representation as  
\begin{equation}
    \begin{gathered}
    \mathbf{X}_{out} = \mathbf{X}_{in} + \text{\textsc{softmax}}_{_L}\left(
    (\mathbf{X}_{in} \mathbf{W}_Q) 
    (\mathbf{T}\mathbf{W}_K)^T
    \right) \mathbf{T}, 
    \end{gathered}
    \label{eqn:projector}
\end{equation}
where $\mathbf{X}_{in}, \mathbf{X}_{out} \in \mathbb{R}^{HW\times C}$ are the input and output feature map. $ (\mathbf{X}_{in} \mathbf{W}_Q)   \in \mathbb{R}^{HW\times C}$ is the query computed from the input feature map $\mathbf{X}_{in}$. $(\mathbf{X}_{in} \mathbf{W}_Q)_p \in \mathbb{R}^C$ encodes the information pixel-$p$ requires from the visual tokens. $(\mathbf{T}\mathbf{W}_K) \in \mathbb{R}^{L\times C}$ is the key computed from the token $\mathbf{T}$. $(\mathbf{T}\mathbf{W}_K)_l \in \mathbb{R}^C$ represents the information the $l$-th token encodes. The key-query product determines how to project information encoded in visual tokens $\mathbf{T}$ to the original feature map. $\mathbf{W}_Q \in \mathbb{R}^{C\times C}, \mathbf{W}_K \in \mathbb{R}^{C\times C}$ are learnable weights used to compute queries and keys.

\begin{figure*}[!htb]
\vspace{-0.5cm}
  \centering
  \includegraphics[width=.95\linewidth]{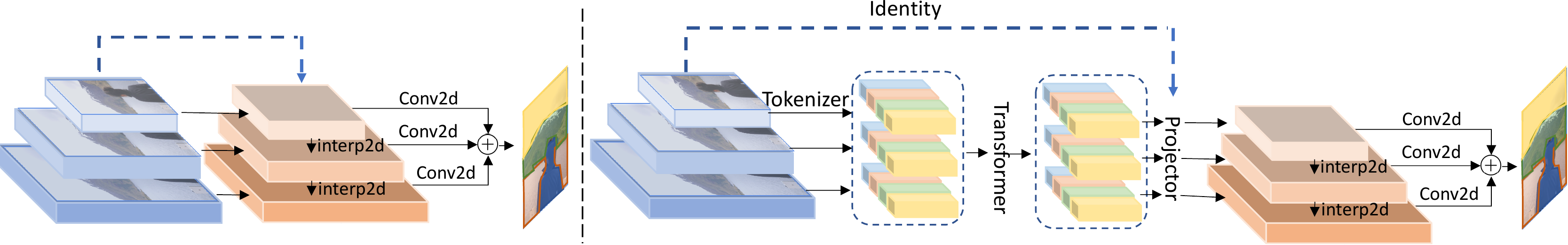}
  \caption{Feature Pyramid Networks (FPN) (left) vs visual-transformer-FPN (VT-FPN) (right) for semantic segmentation. FPN uses convolution and interpolation to merge feature maps with different resolutions. VT-FPN extraxt visual tokens from all feature maps, merge them with one transformer, and project back to the original feature maps.}
  \label{fig:fpn}
\end{figure*}

\section{Using Visual Transformers in vision models}
\label{sec:vt_in_vision_models}
In this section, we discuss how to use VTs as building blocks in vision models. We define three hyper-parameters for each VT: channel size of the feature map; channel size of the visual tokens; and the number of visual tokens. 

\textbf{Image classification model:} For image classification, following the convention of previous work, we build our networks with backbones inherited from ResNet \cite{he2016deep}. Based on ResNet-\{18, 34, 50, 101\}, we build corresponding visual-transformer-ResNets (VT-ResNets) by replacing the last stage of convolutions with VT modules. The last stage of ResNet-\{18, 34, 50, 101\} contains 2 basic blocks, 3 basic blocks, 3 bottleneck blocks, and 3 bottleneck blocks, respectively. We replace them with the same number (2, 3, 3, 3) of VT modules. At the end of stage-4 (before stage-5 max pooling), ResNet-\{18, 34\} generate feature maps with the shape of $14^2 \times 256$, and ResNet-\{50, 101\} generate feature maps with the shape of $14^2 \times 1024$. We set VT's feature map channel size to be 256, 256, 1024, 1024 for ResNet-\{18, 34, 50, 101\}. We adopt 16 visual tokens with a channel size of 1024 for all the models. At the end of the network, we output 16 visual tokens to the classification head, which applies an average pooling over the tokens and use a fully-connected layer to predict the probabilities. A table summarizing the stage-wise description of the model is provided in Appendix \ref{sec:vt-resnet-description}. Since VTs only operate on 16 visual tokens, we can reduce the last stage's FLOPs by up to 6.9x, as shown in Table \ref{tab:reduction}. 
\begin{table}[h]
\vspace{-0.4cm}
\centering

\resizebox{0.4\textwidth}{!}{
 
\begin{tabular}{cc|cccc}
\hline
                                             &         & R18   & R34   & R50   & R101  \\ \hline
\multicolumn{1}{c}{\multirow{2}{*}{FLOPs}}   & Total   & 1.14x & 1.16x & 1.20x & 1.09x \\
\multicolumn{1}{c}{}                        & Stage-5 & 2.4x  & 5.0x  & 6.1x  & 6.9x  \\ \hline
\multicolumn{1}{c}{\multirow{2}{*}{Params}} & Total   & 0.91x & 1.21x & 1.19x & 1.19x \\
\multicolumn{1}{c}{}                        & Stage-5 & 0.9x  & 1.5x  & 1.26x & 1.26x \\ \hline
\end{tabular}

}

\caption{FLOPs and parameter size reduction of VTs on ResNets by replacing the last stage of convolution modules with VT modules.}
\label{tab:reduction}
\end{table}

\textbf{Semantic segmentation}:
We show that using VTs for semantic segmentation can tackle several challenges with the pixel-convolution paradigm. First, the computational complexity of convolution grows with the image resolution. Second, convolutions struggles to capture long-term interactions between pixels. VTs, on the other hand, operate on a small number of visual tokens regardless of the image resolution, and since it models concept interactions in the token-space, it bypasses the ``long-range'' challenge with pixel-arrays. 

To validate our hypothesis, we use panoptic feature pyramid networks (FPN) \cite{kirillov2019panoptic} as a baseline and use VTs to improve the network. 
Panoptic FPNs use ResNet as backbone to extract feature maps from different stages with various resolutions. These feature maps are then fused by a feature pyramid network in a top-down manner to generate a multi-scale and detail preserving feature map with rich semantics for segmentation (Figure \ref{fig:fpn} left). FPN is computationally expensive since it heavily relies on spatial convolutions operating on high resolution feature maps with large channel sizes. We use VTs to replace convolutions in FPN. We name the new module as VT-FPN (Figure \ref{fig:fpn} right). From each resolution's feature map, VT-FPN extract 8 visual tokens with a channel size of 1024. The visual tokens are combined and fed into one transformer to compute interactions between visual tokens across resolutions. The output tokens are then projected back to the original feature maps, which are then used to perform pixel-level prediction. Compared with the original FPN, the computational cost for VT-FPN is much smaller since we only operate on a very small number of visual tokens rather than all the pixels. 
Our experiment shows VT-FPN uses 6.4x fewer FLOPs than FPN while preserving or surpassing its performance (Table \ref{tab:seg_coco} \& \ref{tab:seg_lip}). 

\section{Experiments}
We conduct experiments with VTs on image classification and semantic segmentation to (a) understand the key components of VTs and (b) validate their effectiveness. 

\subsection{Visual Transformer for Classification}
\label{sec:ablation}
We conduct experiments on the ImageNet dataset \cite{deng2009imagenet} with around 1.3 million images in the training set and 50 thousand images in the validation set. We implement VT models in PyTorch \cite{pytorch}. We use stochastic gradient descent (SGD) optimizer with Nesterov momentum \cite{sutskever2013importance}. We use an initial learning rate of $0.1$, a momentum of $0.9$, and a weight decay of 4e-5. We train the model for 90 epochs, and decay the learning rate by 10x every 30 epochs. We use a batch size of 256 and 8 V100 GPUs for training. 

\textbf{VT vs. ResNet with default training recipe}: In Table \ref{tab:vt-vs-resnet}, we first compare VT-ResNets and vanilla ResNets under the same training recipe. VT-ResNets in this experiment use a filter-based tokenizer for the first VT module and recurrent tokenizers in later modules. We can see that after replacing the last stage of convolutions in ResNet18 and ResNet34, VT-based ResNets use many fewer FLOPs: 244M fewer FLOPs for ResNet18 and 384M fewer for ResNet34. Meanwhile, VT-ResNets achieve much higher top-1 validation accuracy than the corresponding ResNets by up to 2.2 points. This confirms effectiveness of VTs. Also note that the training accuracy achieved by VT-ResNets are much higher than that of baseline ResNets: VT-R18 is 7.9 points higher and VT-R34 is 6.9 points higher. This indicates that VT-ResNets are overfitting more heavily than regular ResNets. We hypothesize this is because VT-ResNets have much larger capacity and we need stronger regularization (e.g., data augmentation) to fully utilize the model capacity. We address this in Section \ref{sec:advanced_recipe} and Table \ref{tab:imagenet}. 

\begin{table}[]
\centering

\resizebox{0.4\textwidth}{!}{

\begin{tabular}{lcccc}
\hline
       & \begin{tabular}[c]{@{}c@{}}Top-1\\ Acc (\%)\\ (Val)\end{tabular} & \begin{tabular}[c]{@{}c@{}}Top-1\\ Acc (\%)\\ (Train)\end{tabular} & \begin{tabular}[c]{@{}c@{}}FLOPs\\ (M)\end{tabular} & \begin{tabular}[c]{@{}c@{}}Params\\ (M)\end{tabular} \\ \hline
R18    & 69.9                                                             & 68.6                                                               & 1814                                                & 11.7                                                 \\
VT-R18 & \textbf{72.1}                                                    & 76.5                                                               & 1570                                                & 11.7                                                 \\ \hline
R34    & 73.3                                                             & 73.9                                                               & 3664                                                & 21.8                                                 \\
VT-R34 & \textbf{75.0}                                                    & 80.8                                                               & 3280                                                & 21.9                                                 \\ \hline
\end{tabular}

}

\caption{VT-ResNet vs. baseline ResNets on the ImageNet dataset. By replacing the last stage of ResNets, VT-ResNet uses 224M, 384M fewer FLOPs than the baseline ResNets while achieving 1.7 points and 2.2 points higher validation accuracy. Note the training accuracy of VT-ResNets are much higher. This indicates VT-ResNets have higher model capacity and require stronger regularization (e.g., data augmentation) to fully utilize the model. See Table \ref{tab:imagenet}.}
\label{tab:vt-vs-resnet}
\end{table}

\textbf{Tokenizer ablation studies}: In Table \ref{tab:tokenizer}, we compare different types of tokenizers used by VTs. We consider a \textit{pooling-based} tokenizer, a \textit{clustering-based} tokenizer, and a filter-based tokenizer (Section \ref{sec:static_tokenization}). We use the candidate tokenizer in the first VT module and use recurrent tokenizers in later modules. As a baseline, we implement a \textit{pooling-based} tokenizer, which spatially downsamples a feature map $\mathbf{X}$ to reduce its spatial dimensions from $HW=196$ to $L=16$, instead of grouping pixels by their semantics. As a more advanced baseline, we consider a clustering-based tokenizer, which is described in Appendix \ref{sec:clustering_based_tokenizer}. It applies K-Means clustering in the semantic space to group pixels to visual tokens. As can be seen from  Table \ref{tab:tokenizer}, filter-based and clustering-based tokenizers perform significantly better than the pooling-based baseline, validating our hypothesis that feature maps contain redundancies, and this can be addressed by grouping pixels in semantic space. The difference between filter-based and clustering-based tokenizers is small and vary between R18 and R34. We hypothesize this is because both tokenizers have their own drawbacks. The filter-based tokenizer relies on fixed convolution filters to detect and assign pixels to semantic groups, and is limited by the capacity of the convolution filters to deal with diverse and sparse high-level semantic concepts. On the other hand, the clustering-based tokenizer extracts semantic concepts that exist in the image, but it is not designed to capture the essential semantic concepts. 

\begin{table}[]
\centering

\resizebox{0.42\textwidth}{!}{

\begin{tabular}{ll|ccc}
\hline
                     &                  & \begin{tabular}[c]{@{}c@{}}Top-1\\ Acc (\%)\end{tabular} & \begin{tabular}[c]{@{}c@{}}FLOPs\\ (M)\end{tabular} & \begin{tabular}[c]{@{}c@{}}Params\\ (M)\end{tabular} \\ \hline
\multirow{3}{*}{R18} & Pooling-based    & 70.5                                                     & 1549                                                & 11.0                                                 \\
                     & Clustering-based & 71.8                                                     & 1579                                                & 11.6                                                 \\ 
                     & Filter-based     & \textbf{72.1}                                            & 1580                                                & 11.7                                                 \\\hline
\multirow{3}{*}{R34} & Pooling-based    & 73.6                                                     & 3246                                                & 20.6                                                 \\
                     & Clustering-based & \textbf{75.2}                                            & 3299                                                & 21.8                                                 \\ 
                     & Filter-based     & 74.9                                                     & 3280                                                & 21.9                                                 \\\hline
\end{tabular}

}

\caption{VT-ResNets using with different types of tokenizers. Pooling-based tokenizers spatially downsample a feature map to obtain visual tokens. Clustering-based tokenizer (Appendix \ref{sec:clustering_based_tokenizer}) groups pixels in the semantic space. Filter-based tokenizers (\ref{sec:static_tokenization}) use convolution filters to group pixels. Both filter-based and cluster-based tokenizers work much better than pooling-based tokenizers, validating the importance of grouping pixels by their semantics. }
\label{tab:tokenizer}
\end{table}

In Table \ref{tab:token_refining}, we validate the recurrent tokenizer's effectiveness. We use a filter-based tokenizer in the first VT module and use recurrent tokenizers in subsequent modules. Experiments show that using recurrent tokenizers leads to higher accuracy.

\begin{table}[t!]
\centering

\vspace{-0.4cm}

\resizebox{0.37\textwidth}{!}{

\begin{tabular}{ll|ccc}
\hline
                     &        & \begin{tabular}[c]{@{}c@{}}Top-1\\ Acc (\%)\end{tabular} & \begin{tabular}[c]{@{}c@{}}FLOPs\\ (M)\end{tabular} & \begin{tabular}[c]{@{}c@{}}Params\\ (M)\end{tabular} \\ \hline
\multirow{2}{*}{R18} & w/ RT  & \textbf{72.0}                                            & 1569                                                & 11.7                                                 \\
                     & w/o RT & 71.2                                                     & 1586                                                & 11.1                                                 \\ \hline
\multirow{2}{*}{R34} & w/ RT  & \textbf{74.9}                                            & 3246                                                & 21.9                                                 \\
                     & w/o RT & 74.5                                                     & 3335                                                & 20.9                                                 \\ \hline
\end{tabular}

}

\caption{VT-ResNets that use recurrent tokenizers achieve better performance, since recurrent tokenizers are content-aware. RT denotes recurrent tokenizer.}
\label{tab:token_refining}
\end{table}

\textbf{Modeling token relationships}: In Table \ref{tab:token_interaction}, we compare different methods of capturing token relationships. As a baseline, we do not compute the interactions between tokens. This leads to the worst performance among all variations. This validates the necessities of capturing the relationship between different semantic concepts. Another alternative is to use graph-convolutions similar to \cite{chen2019graph,liang2018symbolic,zhang2019latentgnn}, but its performance is worse than that of VTs. This is likely due to the fact that graph-convolutions bind each visual token to a fixed semantic concept. In comparison, using transformers allows each visual token to encode any semantic concepts as long as the concept appears in the image.  This allows the models to fully utilize its capacity. 

\begin{table}[t!]
\centering

\resizebox{0.39\textwidth}{!}{

\begin{tabular}{ll|ccc}
\hline
                     &             & \begin{tabular}[c]{@{}c@{}}Top-1\\ Acc (\%)\end{tabular} & \begin{tabular}[c]{@{}c@{}}FLOPs\\ (M)\end{tabular} & \begin{tabular}[c]{@{}c@{}}Params\\ (M)\end{tabular} \\ \hline
\multirow{3}{*}{R18} & None        & 68.7                                                     & 1528                                                & 8.5                                                  \\
                     & GraphConv   & 69.3                                                     & 1528                                                & 8.5                                                  \\
                     & Transformer & \textbf{71.5}                                            & 1580                                                & 11.7                                                 \\ \hline
\multirow{3}{*}{R34} & None        & 73.3                                                     & 3222                                                & 17.1                                                 \\
                     & GraphConv   & 73.7                                                     & 3223                                                & 17.1                                                 \\
                     & Transformer & \textbf{75.2}                                            & 3299                                                & 21.8                                                 \\ \hline
\end{tabular}

}

\caption{VT-ResNets using different modules to model token relationships. Models using transformers perform better than graph-convolution or no token-space operations. This validates that it is important to model relationships between visual token (semantic concepts) and transformer work better than graph convolution in relating tokens.}
\label{tab:token_interaction}
\vspace{-0.3cm}
\end{table}

\textbf{Token efficiency ablation}: In Table \ref{tab:num_tokens}, we test varying numbers of visual tokens, only to find negligible or no increase in accuracy. This agrees with our hypothesis that VTs are already capturing a wide variety of concepts with just a few handfuls of tokens--additional tokens are not needed, as the space of possible, high-level concepts is already covered. %\wbc{Alvin check here. }

\textbf{Pojection ablation}: In Table \ref{tab:proj}, we study whether we need to project visual tokens back to feature maps. We hypothesize that projecting the visual tokens is an important step since in vision understanding, pixel-level semantics are very important, and visual tokens are representations in the semantic space that do not encode any spatial information. As validated by Table \ref{tab:proj}, projecting visual tokens back to the feature map leads to higher performance, even for image classification tasks.

\begin{table}[t!]
\vspace{-0.4cm}

\centering

\resizebox{0.35\textwidth}{!}{

\begin{tabular}{cc|ccc}
\hline
                     & \begin{tabular}[c]{@{}c@{}}No.\\ Tokens\end{tabular} & \begin{tabular}[c]{@{}c@{}}Top-1\\ Acc (\%)\end{tabular} & \begin{tabular}[c]{@{}c@{}}FLOPs\\ (M)\end{tabular} & \begin{tabular}[c]{@{}c@{}}Params\\ (M)\end{tabular} \\ \hline
\multirow{3}{*}{R18} & 16                                                   & 71.8                                                     & 1579                                                & 11.6                                                 \\
                     & 32                                                   & 71.9                                                     & 1711                                                & 11.6                                                 \\
                     & 64                                                   & \textbf{72.1}                                            & 1979                                                & 11.6                                                 \\ \hline
\multirow{3}{*}{R34} & 16                                                   & \textbf{75.1}                                            & 3299                                                & 21.8                                                 \\
                     & 32                                                   & 75.0                                                     & 3514                                                & 21.8                                                 \\
                     & 64                                                   & 75.0                                                     & 3952                                                & 21.8                                                 \\ \hline
\end{tabular}

}

\caption{Using more visual tokens do not improve the accuracy of VT, which agrees with our hypothesis that images can be described by a compact set of visual tokens.}
\label{tab:num_tokens}
\end{table}

\begin{table}[]
\centering

\resizebox{0.4\textwidth}{!}{

\begin{tabular}{ll|ccc}
\hline
                     &               & \begin{tabular}[c]{@{}c@{}}Top-1\\ Acc (\%)\end{tabular} & \begin{tabular}[c]{@{}c@{}}FLOPs\\ (M)\end{tabular} & \begin{tabular}[c]{@{}c@{}}Params\\ (M)\end{tabular} \\ \hline
\multirow{2}{*}{R18} & w/ projector  & \textbf{72.0}                                            & 1569                                                & 11.7                                                 \\
                     & w/o projector & 71.0                                                     & 1498                                                & 9.4                                                  \\ \hline
\multirow{2}{*}{R34} & w/ projector  & \textbf{74.8}                                            & 3280                                                & 21.9                                                 \\
                     & w/o projector & 73.9                                                     & 3159                                                & 17.4                                                 \\ \hline
\end{tabular}

}

\caption{VTs that projects tokens back to feature maps perform better. This may be because feature maps still encode important spatial information.}
\label{tab:proj}
\vspace{-0.3cm}
\end{table}

\subsection{Training with Advanced Recipe}
\label{sec:advanced_recipe}
In Table \ref{tab:vt-vs-resnet}, we show that under the regular training recipe, the VT-ResNets experience serious overfitting. Despite their accuracy improvement on the validation set, its training accuracy improves by a significantly larger margin. We hypothesize that this is because VT-based models have much higher model capacity. To fully unleash the potential of VT, we used a more advanced training recipe to train VT models. To prevent overfitting, we train with more training epochs, stronger data augmentation, stronger regularization, and distillation. Specifically, we train VT-ResNet models for 400 epochs with RMSProp. We use an initial learning rate of 0.01 and increase to 0.16 in 5 warmup epochs, then reduce the learning rate by a factor of 0.9875 per epoch. We use synchronized batch normalization and distributed training with a batch size of 2048. We use label smoothing and AutoAugment \cite{cubuk2019autoaugment} and we set the stochastic depth survival probability~\cite{huang2016deep} and dropout ratio to be 0.9 and 0.2, respectively.  We use exponential moving average (EMA) on the model weights with 0.99985 decay. We use knowledge distillation~\cite{hinton2015distilling} in the training recipe, where the teacher model is FBNetV3-G~\cite{dai2020fbnetv3}. The total loss is a weighted sum of distillation loss ($\times$0.8) and cross entropy loss ($\times$0.2). 

Our results are reported in Table \ref{tab:imagenet}. Compared with the baseline ResNet models, VT-ResNet models achieve 4.6 to 7 points higher accuracy and surpass all other related work that adopt attention of different forms based on ResNets \cite{hu2018squeeze,woo2018cbam,bello2019attention, chen20182,hu2019local, ramachandran2019stand,zhao2020exploring,chen2019graph}. This validates that our advanced training recipe better utilizes the model capacity of VT-ResNet models to outperform all previous baselines.

Note that in addition to the architecture differences, previous works also used their own training recipes and it is infeasible for us to try these recipes one by one. So to understand the source of the accuracy gain, we use the same training recipe to train baseline ResNet18 and ResNet34 and also observe significant accuracy improvement on baseline ResNets. But note that under the advanced training recipe, the accuracy gap between VT-ResNet and baselines increases from 1.7 and 2.2 points to 2.2 and 3.0 points. This further validates that a stronger training recipe can better utilize the model capacity of VTs. While achieving higher accuracy than previous work, VT-ResNets also use much fewer FLOPs and parameters, even we only replace the last stage of the baseline model. If we consider the reduction over the original stage, we observe FLOP reductions of up to 6.9x, as shown in Table \ref{tab:reduction}. 

\begin{table}[t!]
\vspace{-0.4cm}

\centering

\resizebox{0.38\textwidth}{!}{

\begin{tabular}{r|ccc}
\hline
Models                                    & \begin{tabular}[c]{@{}c@{}}Top-1\\ Acc (\%)\end{tabular} & \begin{tabular}[c]{@{}c@{}}FLOPs\\ (G)\end{tabular} & \begin{tabular}[c]{@{}c@{}}Params\\ (M)\end{tabular} \\ \hline
R18\cite{he2016deep}                      & 69.8                                                     & 1.814                                               & 11.7                                                 \\
R18+SE\cite{hu2018squeeze, woo2018cbam}   & 70.6                                                     & 1.814                                               & 11.8                                                 \\
R18+CBAM\cite{woo2018cbam}                & 70.7                                                     & 1.815                                               & 11.8                                                 \\
LR-R18\cite{hu2019local}                  & 74.6                                                     & 2.5                                                 & 14.4                                                 \\
R18\cite{he2016deep}(ours)                & 73.8                                                     & 1.814                                               & 11.7                                                 \\
VT-R18(ours)                              & \textbf{76.8}                                            & \textbf{1.569}                                      & \textbf{11.7}                                        \\ \hline
R34\cite{he2016deep}                      & 73.3                                                     & 3.664                                               & 21.8                                                 \\
R34+SE\cite{hu2018squeeze, woo2018cbam}   & 73.9                                                     & 3.664                                               & 22.0                                                 \\
R34+CBAM\cite{woo2018cbam}                & 74.0                                                     & 3.664                                               & 22.9                                                 \\
AA-R34\cite{bello2019attention}           & 74.7                                                     & 3.55                                                & 20.7                                                 \\
R34\cite{he2016deep}(ours)                & 77.7                                                     & 3.664                                               & 21.8                                                 \\
VT-R34(ours)                              & \textbf{79.9}                                            & \textbf{3.236}                                      & \textbf{19.2}                                        \\ \hline
R50\cite{he2016deep}                      & 76.0                                                     & 4.089                                               & 25.5                                                 \\
R50+SE\cite{hu2018squeeze, woo2018cbam}   & 76.9                                                     & 3.860*                                              & 28.1                                                 \\
R50+CBAM\cite{woo2018cbam}                & 77.3                                                     & 3.864*                                              & 28.1                                                 \\
LR-R50\cite{hu2019local}                  & 77.3                                                     & 4.3                                                 & 23.3                                                 \\
Stand-Alone\cite{ramachandran2019stand}   & 77.6                                                     & 3.6                                                 & \textbf{18.0}                                        \\
AA-R50\cite{bello2019attention}           & 77.7                                                     & 4.1                                                 & 25.6                                                 \\
$A^2$-R50\cite{chen20182}                 & 77.0                                                     & -                                                   & -                                                    \\
SAN19\cite{zhao2020exploring}             & 78.2                                                     & \textbf{3.3}                                        & 20.5                                                 \\
GloRe-R50\cite{chen2019graph}             & 78.4                                                     & 5.2                                                 & 30.5                                                 \\
VT-R50(ours)                              & \textbf{80.6}                                            & 3.412                                               & 21.4                                                 \\ \hline
R101\cite{he2016deep}                     & 77.4                                                     & 7.802                                               & 44.4                                                 \\
R101+SE \cite{hu2018squeeze, woo2018cbam} & 77.7                                                     & 7.575*                                              & 49.3                                                 \\
R101+CBAM\cite{woo2018cbam}               & 78.5                                                     & 7.581*                                              & 49.3                                                 \\
LR-R101\cite{hu2019local}                 & 78.5                                                     & 7.79                                                & 42.0                                                 \\
AA-R101\cite{bello2019attention}          & 78.7                                                     & 8.05                                                & 45.4                                                 \\
GloRe-R200\cite{chen2019graph}            & 79.9                                                     & 16.9                                                & 70.6                                                 \\
VT-R101(ours)                             & \textbf{82.3}                                            & \textbf{7.129}                                      & \textbf{41.5}                                        \\ \hline
\end{tabular}

}
\caption{Comparing VT-ResNets with other attention-augmented ResNets on ImageNet. *The baseline ResNet FLOPs reported in \cite{woo2018cbam} is lower than our baseline.}
\label{tab:imagenet}
\end{table}

\begin{figure*}[h]
  \centering
  \includegraphics[width=.9\linewidth]{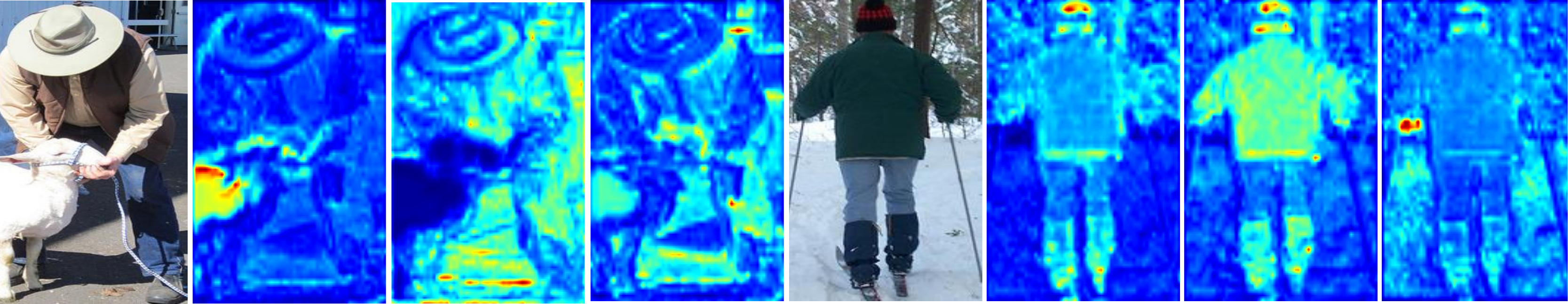}
  \caption{Visualization of the spatial attention generated by a filter-based tokenizer on images from the LIP dataset. Red denotes higher attention values and color blue denotes lower. Without any supervision, visual tokens automatically focus on different areas of the image that correspond to different semantic concepts, such as sheep, ground, clothes, woods.}
  \label{fig:vis}
\end{figure*}

\subsection{Visual Transformer for Semantic Segmentation}
We conduct experiments to test the effectiveness of VT for semantic segmentation on the COCO-stuff dataset \cite{caesar2018cvpr} and the LIP dataset \cite{liang2018look}. The COCO-stuff dataset contains annotations for 91 stuff classes with 118K training images and 5K validation images. LIP dataset is a collection of images containing humans with challenging poses and views. For the COCO-stuff dataset, we train a VT-FPN model with ResNet-\{50, 101\} backbones. Our implementation is based on Detectron2 \cite{wu2019detectron2}. Our training recipe is based on the semantic segmentation FPN recipe with 1x training steps, except that we use synchronized batch normalization in the VT-FPN, change the batch size to 32, and use a base learning rate of 0.04. For the LIP dataset, we also use synchronized batch normalization with a batch size of 96. We optimize the model via SGD with weight decay of 0.0005 and learning rate of 0.01. 

As we can see in Table \ref{tab:seg_coco} and \ref{tab:seg_lip}, after replacing FPN with VT-FPN, both ResNet-50 and ResNet-101 based models achieve slightly higher mIoU, but VT-FPN requires 6.5x fewer FLOPs than a FPN module.

\begin{table}[]
\vspace{-0.4cm}
\centering
\resizebox{0.42\textwidth}{!}{
\begin{tabular}{cc|ccc}
\hline
                       &        & \begin{tabular}[c]{@{}c@{}}mIoU\\ (\%)\end{tabular} & \begin{tabular}[c]{@{}c@{}}Total\\ FLOPs (G)\end{tabular} & \begin{tabular}[c]{@{}c@{}}FPN\\ FLOPs (G)\end{tabular} \\ \hline
\multirow{2}{*}{R-50}  & FPN    & 40.78                                               & 159                                                       & 55.1                                                    \\
                       & VT-FPN & 41.00                                               & 113 (1.41x)                                               & 8.5 (6.48x)                                             \\ \hline
\multirow{2}{*}{R-101} & FPN    & 41.51                                               & 231                                                       & 55.1                                                    \\
                       & VT-FPN & 41.50                                               & 185 (1.25x)                                               & 8.5 (6.48x)                                             \\ \hline
\end{tabular}
}

\caption{Semantic segmentation results on the COCO-stuff validation set. The FLOPs are calculated with a typical input resolution of 800$\times$1216.}
\label{tab:seg_coco}
\end{table}

\begin{table}[]
\centering
\resizebox{0.42\textwidth}{!}{

\begin{tabular}{cc|ccc}
\hline
                      &        & \begin{tabular}[c]{@{}c@{}}mIoU\\ (\%)\end{tabular} & \begin{tabular}[c]{@{}c@{}}Total \\ FLOPs (G)\end{tabular} & \begin{tabular}[c]{@{}c@{}}FPN \\ FLOPs (G)\end{tabular} \\ \hline
\multirow{2}{*}{R50}  & FPN    & 47.04                                               & 37.1                                                       & 12.8                                                     \\
                      & VT-FPN & 47.39                                               & 26.4 (1.41x)                                               & 2.0 (6.40x)                                              \\ \hline
\multirow{2}{*}{R101} & FPN    & 47.35                                               & 54.4                                                       & 12.8                                                     \\
                      & VT-FPN & 47.58                                               & 43.6 (1.25x)                                               & 2.0 (6.40x)                                              \\ \hline
\end{tabular}

}

\caption{Semantic segmentation results on the Look Into Person validation set. The FLOPs are calculated with a typical input resolution of 473$\times$473.}
\label{tab:seg_lip}
\end{table}

\subsection{Visualizing Visual Tokens}
We hypothesize that visual tokens extracted in the VT correspond to different high-level semantics in the image. To better understand this, we visualize the spatial attention $\mathbf{A} \in \mathbb{R}^{HW\times L}$ generated by filter-based tokenizers, where each $\mathbf{A}_{:, l} \in \mathbb{R}^{HW}$ is an attention map to show how each pixel of the image contributes to token-$l$. We plot the attention map in Figure \ref{fig:vis}, and we can see that without any supervision, different visual tokens attend to different semantic concepts in the image, corresponding to parts of the background or foreground objects. More visualization results are provided in Appendix \ref{sec:vis}.

\section{Conclusion}
The convention in computer vision is to represent images as pixel arrays and to apply the \textit{de facto} deep learning operator -- the convolution. In lieu of this, we propose \textit{Visual Transformers} (VTs), as hallmarks of a new computer vision paradigm, learning and relating sparsely-distributed, high-level concepts far more efficiently: Instead of pixel arrays, VTs represent just the high-level concepts in an image using \textit{visual tokens}. Instead of convolutions, VTs apply transformers to directly relate semantic concepts in token-space. To evaluate this idea, we replace convolutional modules with VTs, obtaining significant accuracy improvements across tasks and datasets.
Using an advanced training recipe, our VT improves ResNet accuracy on ImageNet by 4.6 to 7 points. For semantic segmentation on LIP and COCO-stuff, VT-based feature pyramid networks (FPN) achieve 0.35 points higher mIoU despite 6.5x fewer FLOPs than convolutional FPN modules. 
This paradigm can furthermore be compounded with other contemporaneous tricks beyond the scope of this paper, including extra training data and neural architecture search. However, instead of presenting a mosh pit of deep learning tricks, our goal is to show that the pixel-convolution paradigm is fraught with redundancies. To compensate, modern methods add exceptional amounts of computational complexity. However, as model designers and practitioners, we can tackle the root cause instead of exacerbating compute demands, addressing redundancy in the pixel-convolution convention by adopting the newfound token-transformer paradigm moving forward.

\clearpage 

{\small
\bibliographystyle{ieee_fullname}
\bibliography{egbib}
}

\clearpage 

\appendix

\begin{figure}[]
  \centering
  \includegraphics[width=\linewidth]{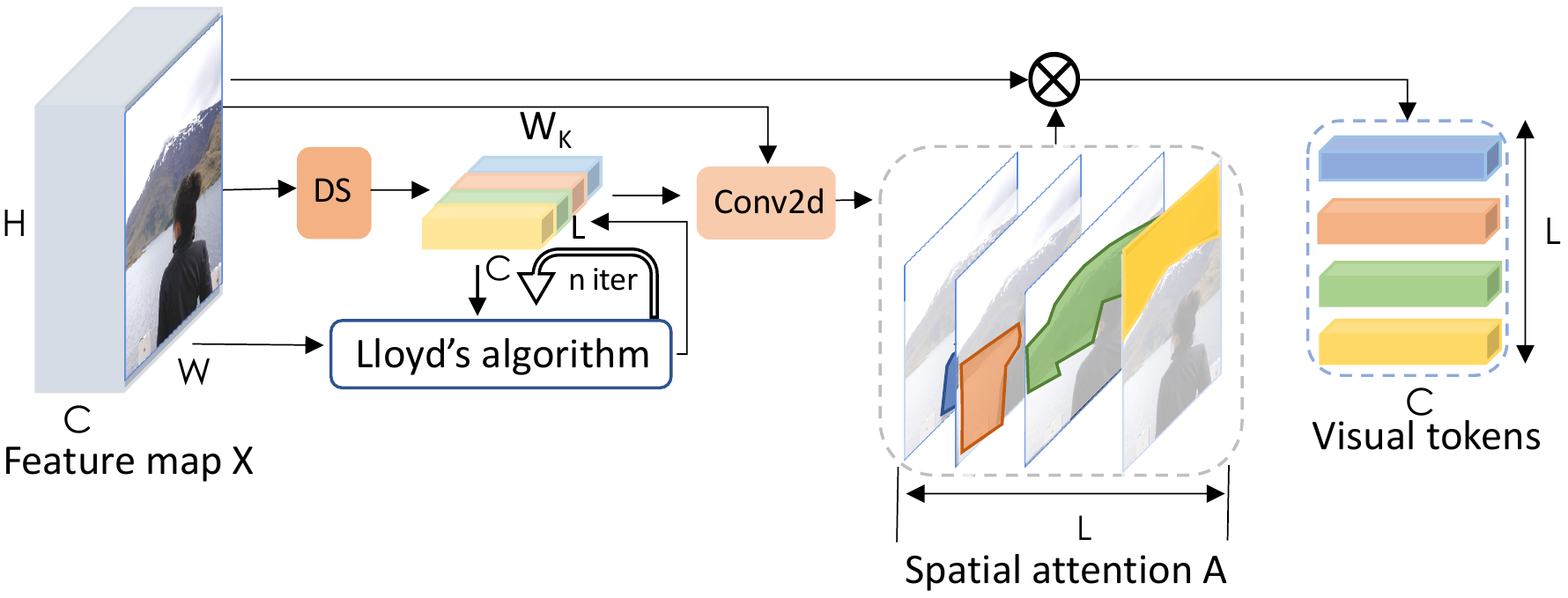}
s  \caption{Cluster-based tokenizer that group pixels using the K-Means centroids of the pixels in the semantic space.}
  \label{fig:cluster_tknzr}
\end{figure}

\begin{table*}[h]
\centering
\begin{tabular}{c|c|cccc}
\hline
Stage & Resolution        & \multicolumn{1}{c|}{VT-R18}                                                                & \multicolumn{1}{c|}{VT-R34}                                                                & \multicolumn{1}{c|}{VT-R50}                                                                 & VT-R101                                                                \\ \hline
1     & 56$\times$56      & \multicolumn{4}{c}{conv7$\times$7-C64-S2 $\rightarrow$ maxpool3$\times$3-S2}                                                                                                                                                                                                                                                                                       \\ \hline
2     & 56$\times$56      & \multicolumn{1}{c|}{BB-C64 $\times2$}                                                       & \multicolumn{1}{c|}{BB-C64 $\times3$}                                                       & \multicolumn{1}{c|}{BN-C256 $\times3$}                                                       & BN-C256 $\times3$                                                       \\ \hline
3     & 28$\times$28      & \multicolumn{1}{c|}{BB-C128 $\times2$}                                                      & \multicolumn{1}{c|}{BB-C128 $\times4$}                                                      & \multicolumn{1}{c|}{BN-C512 $\times4$}                                                       & BN-C512 $\times4$                                                       \\ \hline
4     & 14$\times$14      & \multicolumn{1}{c|}{BB-C256 $\times2$}                                                      & \multicolumn{1}{c|}{BB-C256 $\times6$}                                                      & \multicolumn{1}{c|}{BN-C1024 $\times6$}                                                      & BN-C1024 $\times23$                                                     \\ \hline
5     & 16 & \multicolumn{1}{c|}{\begin{tabular}[c]{@{}c@{}}VT-C512-L16\\ -CT1024 $\times2$\end{tabular}} & \multicolumn{1}{c|}{\begin{tabular}[c]{@{}c@{}}VT-C512-L16\\ -CT1024 $\times3$\end{tabular}} & \multicolumn{1}{c|}{\begin{tabular}[c]{@{}c@{}}VT-C1024-L16\\ -CT1024 $\times3$\end{tabular}} & \begin{tabular}[c]{@{}c@{}}VT-C1024-L16\\ -CT1024 $\times3$\end{tabular} \\ \hline
head  & 1                 & \multicolumn{4}{c}{avgpool-fc1000} 
\\ \hline
\end{tabular}
\caption{Model descriptions for VT-ResNets. VT-R18 denotes visual-transformer-ResNet-18. ``conv7$\times$7-C64-S2'' denotes a 7-by-7 convolution with an output channel size of 64 and a stride of 2. ``BB-C64$\times$2'' denotes a basic block \cite{he2016deep} with an output channel size of 64 and it is repeated twice. ``BN-C256$\times$3'' denotes a bottleneck block  \cite{he2016deep} with an output channel size of 256 and it is repeated by three times.   ``VT-C512-L16-CT1024 $\times$2'' denotes a VT block with a channel size for the output feature map as 512, channel size for visual tokens as 1024, and the number of tokens as 16.
}
\label{tab:VT-R}
\end{table*}

\section{Stage-wise model description of VT-ResNet}
\label{sec:vt-resnet-description}
In this section, we provide a stage-wise description of the model configurations for VT-based ResNet (VT-ResNet). We use three hyper-parameters to control a VT module: channel size of the output feature map C, channel size of visual token CT, and the number of visual tokens L. The model configurations are described in Table \ref{tab:VT-R}.

\section{More visualization results}
\label{sec:vis}
\begin{figure*}[h]
\centering

\includegraphics[width=\linewidth]{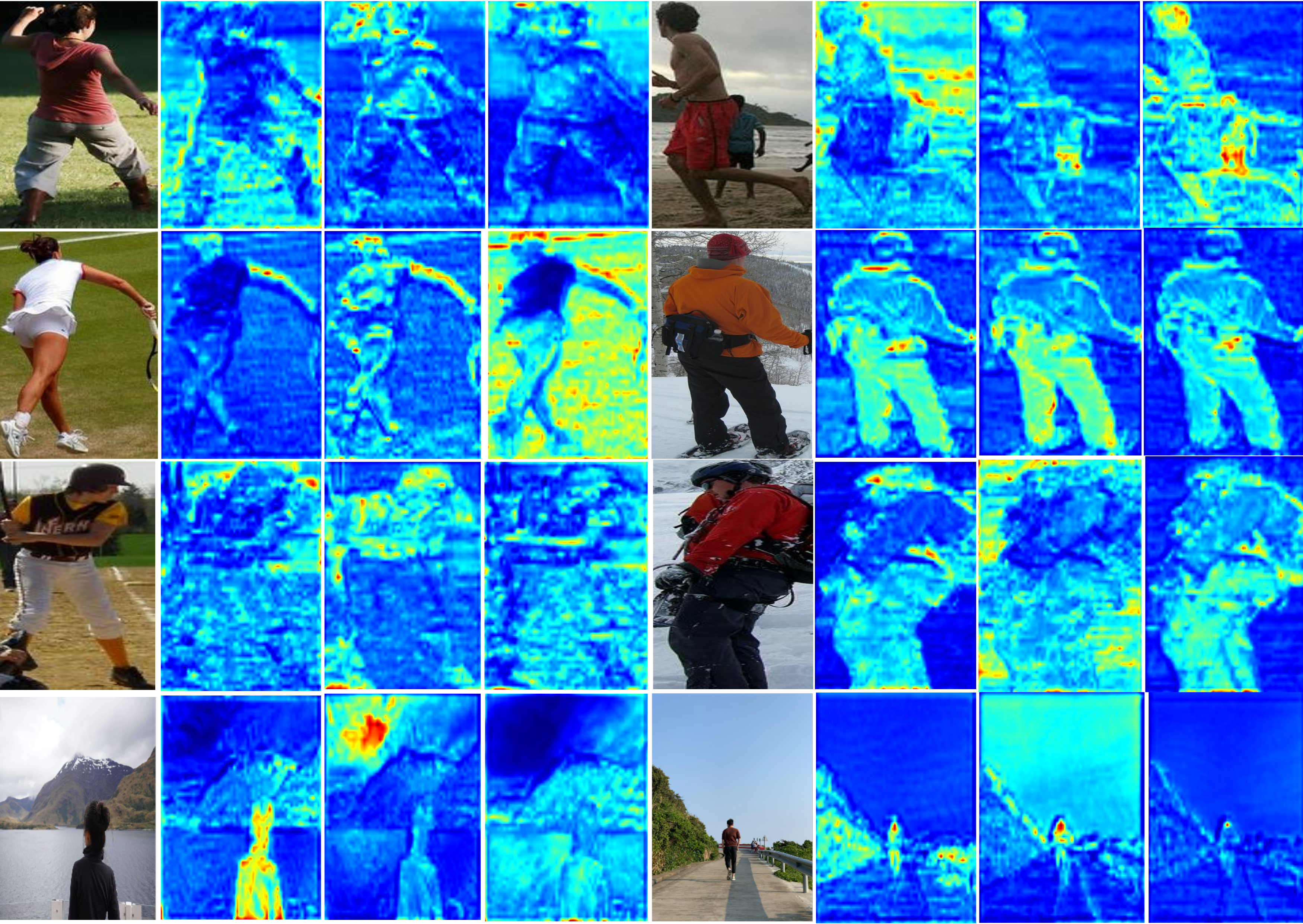}

\caption{Visualization of the spatial attention generated by a filter-based tokenizer. Red denotes higher attention values and color blue denotes lower. Without any supervision, visual tokens automatically focus on different areas of the image that correspond to different semantic concepts.}
\label{fig:lip_supp}
\end{figure*}

We provide more visualization of the spatial attention on images from the LIP dataset in Figure \ref{fig:lip_supp}.

\section{Clustering-based tokenizer}
\label{sec:clustering_based_tokenizer}
To address this limitation of filter-based tokenizers, we devise a content-aware $\mathbf{W}_K$ variant of $\mathbf{W}_A$ to form semantic groups from $\mathbf{X}$. Our insight is to extract concepts present in the current image by clustering pixels, instead of applying the same filters regardless of the image content. First, we treat each pixel as a sample $\{\mathbf{X}_p\}_{p=1}^{HW}$, and apply k-means to find $L$ centroids, which are stacked to form $\mathbf{W}_K \in \mathbb{R}^{C\times L}$. Each centroid represents one semantic concept in the image.  % $\mathbf{U}_l \in \mathbb{R}^C$
Second, we replace $\mathbf{W}_A$ in Equation (\ref{eqn:conv_based_token}) with $\mathbf{W}_K$ to form $L$ semantic groups of channels:
\begin{equation}
    \begin{gathered}
        \mathbf{W}_K = \text{\textsc{kmeans}}(\mathbf{X}), \\
        \mathbf{T} = \text{\textsc{softmax}}_{_{HW}}\left(\mathbf{X}\mathbf{W}_K\right)^T \mathbf{X}.
    \end{gathered}
    \label{eqn:cluster_based_token}
\end{equation}
Pseudo-code for our K-means implementation is provided in Listing \ref{lst:kmeans}, and can be summarized as: Normalize all pixels to unit vectors, initialize centroids with a spatially-downsampled feature map, and run Lloyd's algorithm to produce centroids.

\begin{lstlisting}[language=Python, caption=Pseudo-code of K-Means implemented in PyTorch-like language, label={lst:kmeans}]
def kmeans(X_nchw, L, niter):
  # Input:
  #   X_nchw - feature map
  #   L - num token
  #   niter - num iters of Lloyd's
  N, C, H, W = X_nchw.size()
  # Initialization as down-sampled X
  U_ncl = downsample(X).view(N, C, L)
  X_ncp = X_nchw.view(N, C, H*W)  # p = h*w
  # Normalize to unit vectors
  U_ncl = U_ncl.normalize(dim=1)
  X_ncp = X_ncp.normalize(dim=1)
  for _ in range(niter):  # Lloyd's algorithm
    dist_npl = (
        X_ncp[..., None] - U_ncl[:, :, None, :]
    ).norm(dim=1)
    mask_npl = (dist_npl == dist_npl.min(dim=2)
    U_ncl = X_ncp.MatMul(mask_npl) 
    U_ncl = U_ncl / mask_npl.sum(dim=1)
    U_ncl = U_ncl.normalize(dim=1)
  return U_ncl  # centroids
\end{lstlisting}

Although this tokenizer efficiently models only concepts in the current image, the drawback is that it is not designed to choose the most discriminative concepts.

\end{document}